\documentclass{article} 
\usepackage{iclr2017_conference,times}
\iclrfinalcopy
\usepackage{hyperref}
\usepackage{url}
 
\usepackage[utf8]{inputenc} 
\usepackage[T1]{fontenc}    
\usepackage{booktabs}       
\usepackage{amsfonts}       
\usepackage{nicefrac}       
\usepackage{microtype}      
 
\usepackage{pbox}

\usepackage[english]{babel} 

\usepackage[utf8]{inputenc} 
\usepackage[T1]{fontenc} 

\usepackage{booktabs,array} 

\usepackage{amsmath} 
\usepackage{cancel}
\usepackage{graphicx} 
\usepackage{etoolbox}
\usepackage[norule]{footmisc} 
\usepackage{lastpage} 

\usepackage{tabularx}

\newcommand{\mcX}{\mathcal{X}}
\newcommand{\mcU}{\mathcal{U}}
\newcommand{\mcZ}{\mathcal{Z}}

\newcommand{\eq}[1]{\begin{align*}#1\end{align*}}
\newcommand\numberthis{\addtocounter{equation}{1}\tag{\theequation}}

\usepackage{subfig}
\usepackage{multicol}
\usepackage{pbox}
\usepackage{tikz}
\usepackage{rotating}
\usepackage{hvfloat}

\usepackage{cleveref}

\usepackage[
]{todonotes}

\usepackage{natbib}

\usetikzlibrary{fit,arrows,shapes}
\usetikzlibrary{decorations.pathreplacing}

\let\originalleft\left
\let\originalright\right
\renewcommand{\left}{\mathopen{}\mathclose\bgroup\originalleft}
\renewcommand{\right}{\aftergroup\egroup\originalright}

\newcommand{\RR}{{\mathbb{R}}}

\newcommand{\EE}{{\mathbb{E}}}

\newcommand{\bx}{\ensuremath{\mathbf{x}}}
\newcommand{\xt}{\bx_{t}}
\newcommand{\xtnext}{\bx_{t+1}}
\newcommand{\xts}{\bx_{1:t}}
\newcommand{\xTs}{\bx_{1:T}}

\newcommand{\bw}{\ensuremath{\mathbf{w}}}
\newcommand{\wt}{\bw_{t}}

\newcommand{\wTs}{\bw_{1:T}}

\newcommand{\bv}{\ensuremath{\mathbf{v}}}
\newcommand{\vt}{\bv_{t}}

\newcommand{\vTs}{\bv_{1:T}}

\newcommand{\bz}{\ensuremath{\mathbf{z}}}
\newcommand{\zt}{\bz_{t}}
\newcommand{\ztnext}{\bz_{t+1}}

\newcommand{\zTs}{\bz_{1:T}}

\newcommand{\bu}{\mathbf{u}}
\newcommand{\ut}{\bu_{t}}

\newcommand{\uTs}{\bu_{1:T}}

\newcommand{\bB}{\mathbf{B}}
\newcommand{\Bt}{\bB_{t}}

\newcommand{\bA}{\mathbf{A}}
\newcommand{\At}{\bA_{t}}

\newcommand{\bC}{\mathbf{C}}
\newcommand{\Ct}{\bC_{t}}

\newcommand{\bbeta}{\ensuremath{\boldsymbol{\beta}}}
\newcommand{\betat}{\bbeta_{t}}
\newcommand{\betatnext}{\bbeta_{t+1}}

\newcommand{\betaTs}{\bbeta_{1:T}}

\newcommand{\dint}{\,\mathrm d}

\newcommand{\p}[1]{p\left(#1\right)}
\newcommand{\pgiven}[2]{{\ensuremath{\p{#1\mid#2}}}}
\newcommand{\pundergiven}[3]{{\ensuremath{p_{#2}\left(#1\mid#3\right)}}}

\newcommand{\qundergiven}[3]{{\ensuremath{q_{#2}\left(#1\mid#3\right)}}}

\newcommand{\expectunder}[2]{{\ensuremath{\EE_{#2}\left[#1\right]}}}

\newcommand{\KL}[2]{\ensuremath{\operatorname{KL}\left(#1\mid\mid#2\right)}}

\newcommand{\qphi}{q_\phi}

\title{
Deep Variational Bayes Filters:
Unsupervised Learning of State Space Models from Raw Data
}
\author{Maximilian Karl, Maximilian Soelch, Justin Bayer, Patrick van der Smagt	\\
Data Lab, Volkswagen Group, 80805, M\"unchen, Germany \\
\texttt{\footnotesize zip([maximilian.karl, maximilian.soelch], [@volkswagen.de])}
}

\begin{document}
\maketitle
\begin{abstract}
We introduce Deep Variational Bayes Filters (DVBF), a new method for unsupervised learning and identification of latent Markovian state space models.
Leveraging recent advances in Stochastic Gradient Variational Bayes, DVBF can overcome intractable inference distributions via variational inference. 
Thus, it can handle highly nonlinear  input data with temporal and spatial dependencies such as image sequences without domain knowledge.
Our experiments show that enabling backpropagation through transitions enforces state space assumptions and significantly improves information content of the latent embedding.
This also enables realistic long-term prediction.
\end{abstract}

\section{Introduction}
Estimating probabilistic models for sequential data is central to many domains, such as audio, natural language or physical plants, \cite{graves2013generating,e2c,vrnn2015,deisenroth2011pilco,ko2011learning}.
The goal is to obtain a model $\p{\xTs}$ that best reflects a data set of observed sequences $\xTs$.
Recent advances in {deep learning} have paved the way to powerful models capable of representing high-dimensional sequences with temporal dependencies, e.g., \cite{graves2013generating,e2c,vrnn2015,storn2014}.

Time series for dynamic systems have been studied extensively in systems theory, cf.\ \cite{reviewsystemstheory} and sources therein.
In particular, \emph{state space models} have shown to be a powerful tool to analyze and control the dynamics.
Two tasks remain a significant challenge to this day:
Can we identify the governing system from data only?
And can we perform inference from observables to the latent system variables?
These two tasks are competing:
A more powerful representation of system requires more computationally demanding inference, and efficient inference, such as the well-known Kalman filters, \cite{kalman1961new}, can prohibit sufficiently complex system classes.

Leveraging a recently proposed estimator based on variational inference, stochastic gradient variational Bayes (SGVB, \cite{vae2013, dlgm2014}), approximate inference of latent variables becomes tractable.
Extensions to time series have been shown in \cite{storn2014, vrnn2015}.
Empirically, they showed considerable improvements in marginal data likelihood, i.e., compression, but lack full-information latent states, which prohibits, e.g., long-term sampling.
Yet, in a wide range of applications, full-information latent states should be valued over compression.
This is crucial if the latent spaces are used in downstream applications.

Our contribution is, to our knowledge, the first model that 
(i)~\emph{enforces} the latent state-space model assumptions, allowing for reliable system identification, and plausible long-term prediction of the observable system, 
(ii)~provides the corresponding inference mechanism with rich dependencies, 
(iii)~inherits the merit of neural architectures to be trainable on raw data such as images or other sensory inputs, and 
(iv) scales to large data due to optimization of parameters based on stochastic gradient descent, \cite{bottou2010large}.
Hence, our model has the potential to exploit systems theory methodology for downstream tasks, e.g., control or model-based reinforcement learning, \cite{sutton1996model}.

\section{Background and Related Work}\label{sec:relatedwork}
\subsection{Probabilistic Modeling and Filtering of Dynamical Systems}
We consider non-linear dynamical systems with \emph{observations} $\bx_t\in \mcX \subset \mathbb{R}^{n_x}$, depending on \emph{control inputs} (or \emph{actions})  $\bu_t\in\mcU \subset \mathbb{R}^{n_u}$. 
Elements of $\mcX$ can be high-dimensional sensory data, e.g., raw images.
In particular they may exhibit complex non-Markovian transitions.
Corresponding time-discrete sequences of length T are denoted as $\xTs = (\bx_1, \bx_2, \dots, \bx_T)$ and $\uTs =(\bu_1, \bu_2, \dots, \bu_T)$.

We are interested in a probabilistic model\footnote{Throughout this paper, we consider $\uTs$ as given. The case without any control inputs can be recovered by setting $\mcU = \emptyset$, i.e., not conditioning on control inputs.} $\pgiven{\xTs}{\uTs}$.
Formally, we assume the graphical model
\eq{
	\pgiven{\xTs}{\uTs} = \int \pgiven{\xTs}{\zTs, \uTs}\,\pgiven{\zTs}{\uTs}\dint\zTs,\numberthis\label{eq:graphical_model}	
}
where $\zTs, \,\bz_t\in\mcZ \subset \mathbb{R}^{n_z},$ denotes the corresponding latent sequence.
That is, we assume a generative model with an underlying \emph{latent} dynamical system with \emph{emission model} $\pgiven{\xTs}{\zTs, \uTs}$ and \emph{transition model} $\pgiven{\zTs}{\uTs}$.
We want to learn both components, i.e., we want to perform \emph{latent system identification}.
In order to be able to apply the identified system in downstream tasks, we need to find efficient posterior inference distributions $\pgiven{\zTs}{\xTs}$.
Three common examples are prediction, filtering, and smoothing: inference of $\zt$ from $\bx_{1:t-1}$, $\xts$, or $\xTs$, respectively.
Accurate identification and efficient inference are generally competing tasks, as a wider generative model class typically leads to more difficult or even intractable inference.

The transition model is imperative for achieving good long-term results: a bad transition model can lead to divergence of the latent state.
Accordingly, we put special emphasis on it through a Bayesian treatment.
Assuming that the transitions may differ for each time step, we impose a regularizing prior distribution on a set of \emph{transition parameters} $\betaTs$:
\eq{
	\eqref{eq:graphical_model} = \iint \pgiven{\xTs}{\zTs, \uTs}\pgiven{\zTs}{\betaTs, \uTs}\,\p{\betaTs}\dint\betaTs\dint\zTs\numberthis\label{eq:betas_introduced}
}
To obtain state-space models, we impose assumptions on emission and state transition model,
\eq{
	\pgiven{\xTs}{\zTs, \uTs} &= \prod_{t=1}^T \pgiven{\xt}{\zt}\numberthis\label{eq:markov1},\\
	\pgiven{\zTs}{\betaTs, \uTs} &= \prod_{t=0}^{T-1} \pgiven{\ztnext}{\zt, \ut, \betat}\numberthis\label{eq:markov2}.
}
\Cref{eq:markov1,eq:markov2} assume that the current state $\zt$ contains all necessary information about the current observation $\xt$, as well as the next state $\ztnext$ (given the current control input $\ut$ and transition parameters $\betat$). 
That is, in contrast to observations, $\bz_t$ exhibits Markovian behavior.

A typical example of these assumptions are Linear Gaussian Models (LGMs), i.e., both state transition and  emission model are affine transformations with Gaussian offset noise,
\eq{\ztnext &= \mathbf{F}_t\zt + \mathbf{B}_t\ut + \mathbf{w}_t&\mathbf{w}_t \sim \mathcal{N}(\mathbf{0},\mathbf{Q}_t),\label{eq:kf_transition}\numberthis\\
\xt &= \mathbf{H}_t\zt + \mathbf{y}_t&\mathbf{y}_t \sim \mathcal{N}(\mathbf{0},\mathbf{R}_t).\label{eq:kf_observation}\numberthis}
Typically, \emph{state transition matrix} $\mathbf{F}_t$ and \emph{control-input matrix} $\Bt$ are assumed to be given, so that $\betat = \wt$. \Cref{sub:llt} will show that our approach allows other variants such as $\betat = (\mathbf{F}_t, \mathbf{B}_t, \wt)$.
Under the strong assumptions \eqref{eq:kf_transition} and \eqref{eq:kf_observation} of LGMs, inference is provably solved optimally by the well-known Kalman filters. 
While extensions of Kalman filters to nonlinear dynamical systems exist, \cite{julier1997new}, and are successfully applied in many areas, they suffer from two major drawbacks: firstly, its assumptions are restrictive and are violated in practical applications, leading to suboptimal results. Secondly, parameters such as $\mathbf{F}_t$ and $\Bt$ have to be known in order to perform posterior inference. There have been efforts to learn such system dynamics, cf.\ \cite{ghahramani1996parameter,honkela2010approximate} based on the expectation maximization (EM) algorithm or \cite{valpola2002unsupervised}, which uses neural networks. However, these algorithms are not applicable in cases where the true posterior distribution is intractable. This is the case if, e.g., image sequences are used, since the posterior is then highly nonlinear---typical mean-field assumptions on the approximate posterior are too simplified. Our new approach will tackle both issues, and moreover learn both identification and inference jointly by exploiting Stochastic Gradient Variational Bayes.

\subsection{Stochastic Gradient Variational Bayes (SGVB) for Time Series Distributions}\label{sub:sgvb}
Replacing the bottleneck layer of a deterministic auto-encoder with stochastic units $\bz$, the variational auto-encoder (VAE, \cite{vae2013, dlgm2014}) learns complex marginal data distributions on $\bx$ in an unsupervised fashion from simpler distributions via the graphical model 
\eq{\p{\bx} = \int\p{\bx, \bz}\dint\bz = \int\pundergiven{\bx}{}{\bz}\p{\bz}\dint\bz.} 
In VAEs, $\pgiven{\bx}{\bz} \equiv \pundergiven{\bx}{\theta}{\bz}$ is typically parametrized by a neural network with parameters $\theta$. Within this framework, models are trained by maximizing a lower bound to the marginal data log-likelihood via stochastic gradients:
\eq{
	\ln \p{\bx} \geq \expectunder{\ln\pundergiven{\bx}{\theta}{\bz}}{\qundergiven{\bz}{\phi}{\bx}} - \KL{\qundergiven{\bz}{\phi}{\bx}}{\p{\bz}} =: \mathcal{L}_{\mathrm{SGVB}}(\bx,\phi,\theta)\numberthis\label{eq:loss_VAE}
	}
This is provably equivalent to minimizing the KL-divergence between the \emph{approximate posterior} or \emph{recognition model} ${\qundergiven{\bz}{\phi}{\bx}}$ and the true, but usually intractable posterior distribution $\pgiven{\bz}{\bx}$. $\qphi$ is parametrized by a neural network with parameters $\phi$.

The principle of VAEs has been transferred to time series, \cite{storn2014,vrnn2015}.
Both employ nonlinear state transitions in latent space, but violate \cref{eq:markov2}: 
Observations are directly included in the transition process. 
Empirically, reconstruction and compression work well. The state space $\mcZ$, however, does not reflect all information available, which prohibits plausible generative long-term prediction. Such phenomena with generative models have been explained in \cite{theis2015note}. 

In \cite{deepkalman}, the state-space assumptions \eqref{eq:markov1} and \eqref{eq:markov2} are softly encoded in the Deep Kalman Filter (DKF) model. 
Despite that, experiments, cf.\ \cref{sec:experiments}, show that their model fails to extract information such as velocity (and in general time derivatives), which leads to similar problems with prediction.

\cite{svae} give an algorithm for general graphical model variational inference, not tailored to dynamical systems.
In contrast to previously discussed methods, it does not violate \cref{eq:markov2}. 
The approaches differ in that the recognition model outputs node potentials in combination with message passing to infer the latent state. 
Our approach focuses on learning dynamical systems for control-related tasks and therefore uses a neural network for inferring the latent state directly instead of an inference subroutine.

Others have been specifically interested in applying variational inference for controlled dynamical systems.
In \cite{e2c} (Embed to Control---E2C), a VAE is used to learn the mappings to and from latent space. 
The regularization is clearly motivated by \cref{eq:loss_VAE}. 
Still, it fails to be a mathematically correct lower bound to the marginal data likelihood. 
More significantly, their recognition model requires all observations that contain information w.r.t.\ the current state.
This is nothing short of an additional \emph{temporal} i.i.d.\ assumption on data:
Multiple raw samples need to be stacked into one training sample such that all latent factors (in particular all time derivatives) are present within one sample.
The task is thus greatly simplified, because instead of time-series, we learn a static auto-encoder on the processed data.

A pattern emerges: good prediction should boost compression. 
Still, previous methods empirically excel at compression, while prediction will not work.
We conjecture that this is caused by previous methods trying to fit the latent dynamics to a latent state that is beneficial for \emph{reconstruction}. 
This encourages learning of a stationary auto-encoder with focus of extracting as much from a single observation as possible. 
Importantly, it is not necessary to know the entire sequence for excellent reconstruction of single time steps. 
Once the latent states are set, it is hard to adjust the transition to them. 
This would require changing the latent states slightly, and that comes at a cost of decreasing the reconstruction (temporarily). 
The learning algorithm is stuck in a local optimum with good reconstruction and hence good compression only.
Intriguingly, E2C bypasses this problem with its data augmentation.

This leads to a key contribution of this paper:
We \emph{force the latent space to fit the transition}---reversing the direction, and thus achieving the state-space model assumptions and full information in the latent states.

\section{Deep Variational Bayes Filters}\label{sec:dvbf}

\begin{figure}
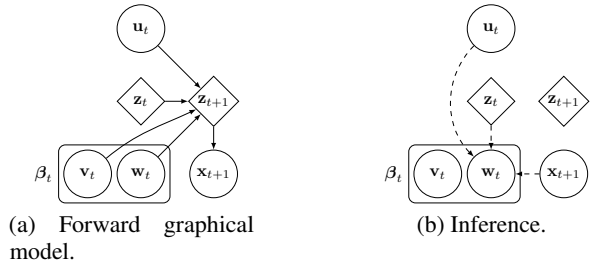
\centering
	\subfloat[Forward graphical model.]{
		\resizebox {.22\linewidth} {!} {
			\input{tikz/gm.tex}
		}
		
	}\qquad\qquad
	\subfloat[Inference.]{
		\resizebox {.22\linewidth} {!} {
			\input{tikz/gm_wo_innov.tex}
		}
	}
	\caption{Left: Graphical model for one transition under state-space model assumptions. The updated latent state $\ztnext$ depends on the previous state $\zt$, control input $\ut$, and transition parameters $\betat$. $\ztnext$ contains all information for generating observation $\xtnext$. Diamond nodes indicate a deterministic dependency on parent nodes. Right: Inference performed during training (or while filtering). Past observations are indirectly used for inference as $\zt$ contains all information about them.}\label{fig:gm}
\end{figure}
\subsection{Reparametrizing the Transition}\label{sub:transition_reparametrization}
The central problem for learning latent states system dynamics is efficient inference of a latent space \emph{that obeys state-space model assumptions}.
If the latter are fulfilled, the latent space \emph{must} contain all information.
Previous approaches emphasized good reconstruction, so that the space only contains information necessary for reconstruction of one time step.
To overcome this, we establish gradient paths through transitions over time so that the transition becomes the driving factor for shaping the latent space, rather than adjusting the transition to the recognition model's latent space.
The key is to prevent the recognition model $\qundergiven{\zTs}{\phi}{\xTs}$ from directly drawing the latent state $\zt$.

Similar to the reparametrization trick from \cite{vae2013, dlgm2014} for making the Monte Carlo estimate differentiable w.r.t.\ the parameters, we make the transition differentiable w.r.t.\ the last state and its parameters:
\eq{
	\ztnext = f\left(\zt,\ut,\betat\right)\numberthis\label{eq:general_transition}
}
Given the stochastic parameters $\betat$, the state transition is deterministic (which in turn means that by marginalizing $\betat$, we still have a stochastic transition). The immediate and crucial consequence is that errors in reconstruction of $\xt$ from $\zt$ are backpropagated directly through time.

This reparametrization has a couple of other important implications: the recognition model no longer infers latent states $\zt$, but transition parameters $\betat$. In particular, the gradient $\partial \ztnext/\partial\zt$ is well-defined from \eqref{eq:general_transition}---gradient information can be backpropagated through the transition. 

This is different from the method used in \cite{deepkalman}, where the transition only occurs in the KL-divergence term of their loss function (a variant of \cref{eq:loss_VAE}). No gradient from the generative model is backpropagated through the transitions.

Much like in \cref{eq:kf_transition}, the stochastic parameters includes a corrective offset term $\wt$, which emphasizes the notion of the recognition model as a filter. In theory, the learning algorithm could still learn the transition as $\ztnext=\wt$. However, the introduction of $\betat$ also enables us to regularize the transition with meaningful priors, which not only prevents overfitting the recognition model, but also enforces meaningful manifolds in the latent space via \emph{transition priors}. Ignoring the potential of the transition over time yields large penalties from these priors. Thus, the problems outlined in \Cref{sec:relatedwork} are overcome by construction.

To install such transition priors, we split $\betat = (\wt,\vt)$. The interpretation of $\wt$ is a sample-specific process noise which can be inferred from incoming data, like in \cref{eq:kf_transition}. On the other hand, $\vt$ are universal transition parameters, which are sample-independent (and are only inferred from data during training). This corresponds to the idea of weight uncertainty in \cite{hinton1993keeping}. This interpretation leads to a natural factorization assumption on the recognition model:
\eq{\qundergiven{\betaTs}{\phi}{\xTs} = \qundergiven{\wTs}{\phi}{\xTs}\,\qphi(\vTs)\numberthis\label{eq:factorizing_recog}}
When using the fully trained model for generative sampling, i.e., sampling without input, the universal state transition parameters can still be drawn from $\qphi(\vTs)$, whereas $\wTs$ is drawn from the prior in the absence of input data. 
\begin{figure}
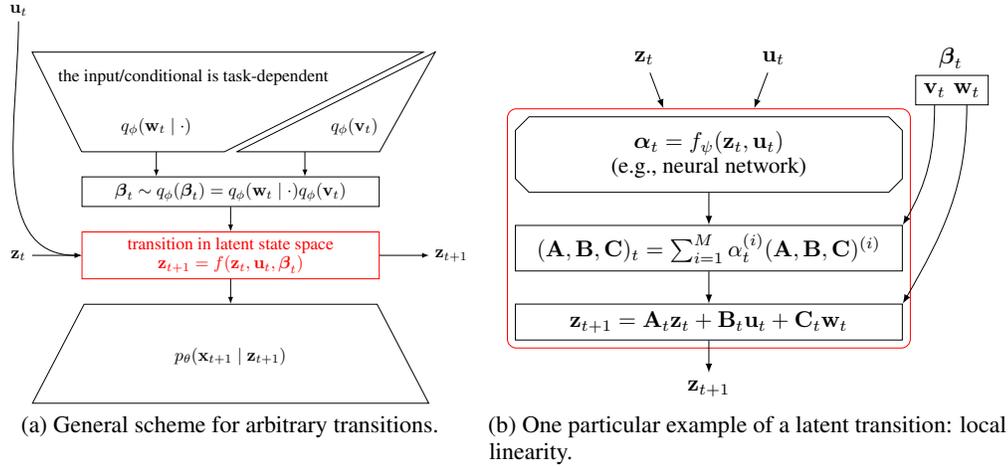

	\centering
	\subfloat[General scheme for arbitrary transitions.]{\label{fig:general_scheme}
		
		\resizebox {.48\linewidth} {!} {
			\input{tikz/scheme.tex}
		}
	}
	\subfloat[One particular example of a latent transition: local linearity.]{\label{fig:llt}
		\resizebox {.48\linewidth} {!} {
			\input{tikz/transition.tex}
		}
	}
	\caption{Left: General architecture for DVBF. Stochastic transition parameters $\betat$ are inferred via the recognition model, e.g., a neural network. Based on a sampled $\betat$, the state transition is computed deterministically. The updated latent state $\ztnext$ is used for predicting $\xtnext$. For details, see \cref{sub:transition_reparametrization}. Right: Zoom into latent space transition (red box in left figure). One exemplary transition is shown, the locally linear transition from \cref{sub:llt}. }
	
\end{figure}

\Cref{fig:gm} shows the underlying graphical model and the inference procedure. \Cref{fig:general_scheme} shows a generic view on our new computational architecture. An example of a locally linear transition parametrization will be given in \cref{sub:llt}.
\subsection{The Lower Bound Objective Function}
In analogy to  \cref{eq:loss_VAE}, we now derive a lower bound to the marginal likelihood $\pgiven{\xTs}{\uTs}$. After reflecting the Markov assumptions \eqref{eq:markov1} and \eqref{eq:markov2} in the factorized likelihood \eqref{eq:betas_introduced}, we have:
\eq{
	\pgiven{\xTs}{\uTs} = \iint \p{\betaTs}\prod_{t=1}^T \pundergiven{\xt}\theta{\zt}\prod_{t=0}^{T-1}\pgiven{\ztnext}{\zt, \ut, \betat}\dint\betaTs\dint\zTs
}
Due to the deterministic transition given $\betatnext$, the last term is a product of Dirac distributions and the overall distribution simplifies greatly:
\eq{
	\pgiven{\xTs}{\uTs}& = \int \p{\betaTs}\prod_{t=1}^T \pundergiven{\xt}\theta{\zt}\Big|_{\zt = f\left(\bz_{t-1},\bu_{t-1},\bbeta_{t-1}\right)}\dint\betaTs\\&\left( = \int \p{\betaTs}\pundergiven{\xTs}\theta{\zTs}\dint\betaTs\right)
}
The last formulation is for notational brevity: the term $\pundergiven{\xTs}\theta{\zTs}$ is \emph{not} independent of $\betaTs$ and $\uTs$. We now derive the objective function, a lower bound to the data likelihood:
\eq{
	\ln\pgiven{\xTs}{\uTs} 
	&= \ln\int\p{\betaTs}\pundergiven{\xTs}\theta{\zTs}\frac{\qundergiven{\betaTs}{\phi}{\xTs, \uTs}}{\qundergiven{\betaTs}{\phi}{\xTs, \uTs}}\dint \betaTs\\
	&\geq \int \qundergiven{\betaTs}{\phi}{\xTs, \uTs}\ln\left(\pundergiven{\xTs}\theta{\zTs}\frac{\p{\betaTs}}{\qundergiven{\betaTs}{\phi}{\xTs, \uTs}}\right)\dint \betaTs\\
	&= \expectunder{\ln\pundergiven{\xTs}\theta{\zTs} - \ln\qundergiven{\betaTs}\phi{\xTs, \uTs} + \ln\p{\betaTs}}{\qphi}\numberthis\label{eq:loss_VIF_pre_anneal}\\
	&= \expectunder{\ln\pundergiven{\xTs}\theta{\zTs}}{\qphi} - \KL{\qundergiven{\betaTs}\phi{\xTs, \uTs}}{\p{\betaTs}}\numberthis\label{eq:loss_VIF}\\
	&=: \mathcal{L}_\mathrm{DVBF}(\xTs,\theta,\phi\mid\uTs)
}
Our experiments show that an annealed version of \eqref{eq:loss_VIF_pre_anneal} is beneficial to the overall performance:
\eq{(\ref{eq:loss_VIF_pre_anneal}') = \expectunder{c_i\ln\pundergiven{\xTs}\theta{\zTs} - \ln\qundergiven{\betaTs}\phi{\xTs, \uTs} + c_i\ln\p{\wTs} + \ln\p{\vTs}}{\qphi}}
Here, $c_i = \max(1, 0.01 + i/T_A)$ is an inverse temperature that increases linearly in the number of gradient updates $i$ until reaching 1 after $T_A$ annealing iterations. Similar annealing schedules have been applied in, e.g., \cite{variationalswitching,variationaltempering,normflow}, where it is shown that they smooth the typically highly non-convex error landscape.
Additionally, the transition prior $p(\vTs)$ was estimated during optimization, i.e., through an empirical Bayes approach. 
In all experiments, we used isotropic Gaussian priors.

\subsection{Example: Locally Linear Transitions}\label{sub:llt}
We have derived a learning algorithm for time series with particular focus on general transitions in latent space. Inspired by \cite{e2c}, this section will show how to learn a particular instance: locally linear state transitions. 
That is, we set \cref{eq:general_transition} to
\eq{
    \ztnext &= \At\zt + \Bt\ut + \Ct \wt, \numberthis \label{eq:first_ass} &t=1,\dots,T,}
where $\wt$ is a stochastic sample from the recognition model and $\At, \Bt,$ and $\Ct$ are matrices of matching dimensions.
They are stochastic functions of $\zt$ and $\ut$ (thus \emph{local} linearity).
We draw \begin{align*}
\vt = \left\{\At^{(i)}, \Bt^{(i)}, \Ct^{(i)}\mid i=1,\dots,M\right\},
\end{align*}
from  $\qphi(\vt)$, i.e., $M$ triplets of matrices, each corresponding to data-\emph{independent}, but learned globally linear system.
These can be learned as point estimates.
We employed a Bayesian treatment as in \cite{weightuncertainty}. 
We yield $\At, \Bt,$ and $\Ct$ as state- and control-\emph{dependent} linear combinations:
\noindent\begin{tabularx}{\linewidth}{@{}XXX@{}}
		\begin{equation*}
			\begin{aligned}
				\hphantom{\boldsymbol{\alpha}_t = f_\psi(\zt, \ut)\in\RR^M}\\
				\At = \sum_{i=1}^{M}\alpha_t^{(i)}\At^{(i)}
			\end{aligned}
		\end{equation*}&
		\begin{equation*}
		\begin{aligned}
		\boldsymbol{\alpha}_t &= f_\psi(\zt, \ut)\in\RR^M\\
		\Bt &= \sum_{i=1}^{M}\alpha_t^{(i)}\Bt^{(i)}
		\end{aligned}
		\end{equation*} &
		\begin{equation*}
			\begin{aligned}
				\hphantom{\boldsymbol{\alpha}_t = f_\psi(\zt, \ut)\in\RR^M}\\
				\Ct = \sum_{i=1}^{M}\alpha_t^{(i)}\Ct^{(i)}
			\end{aligned}
		\end{equation*}
\end{tabularx}
The computation is depicted in \cref{fig:llt}. The function $f_\psi$ can be, e.g., a (deterministic) neural network with weights $\psi$. As a subset of the generative parameters $\theta$, $\psi$ is part of the trainable parameters of our model. The weight vector $\boldsymbol \alpha_t$ is shared between the three matrices.
There is a correspondence to \cref{eq:kf_transition}: $\At$ and $\mathbf{F}_t$, $\Bt$ and $\Bt$, as well as $\Ct\Ct^\top$ and $\mathbf{Q}_t$ are related.

We used this parametrization of the state transition model for our experiments.
It is important that the parametrization is up to the user and the respective application.

\section{Experiments and Results}\label{sec:experiments}
In this section we validate that DVBF with locally linear transitions (DVBF-LL) (\cref{sub:llt}) outperforms Deep Kalman Filters (DKF, \cite{deepkalman}) in recovering latent spaces with full information.
\footnote{We do not include E2C, \cite{e2c}, due to the need for data modification and its inability to provide a correct lower bound as mentioned in \cref{sub:sgvb}.}
We focus on environments that can be simulated with full knowledge of the ground truth latent dynamical system.
The experimental setup is described in the Supplementary Material.
We published the code for DVBF and a link will be made available at \url{https://brml.org/projects/dvbf}.

\subsection{Dynamic Pendulum}
\begin{figure}
	\centering
	\begin{tabular}{c|c}
		\subfloat[DVBF-LL]{
			\begin{tabular}{cc}
				\includegraphics[width=.2\linewidth]{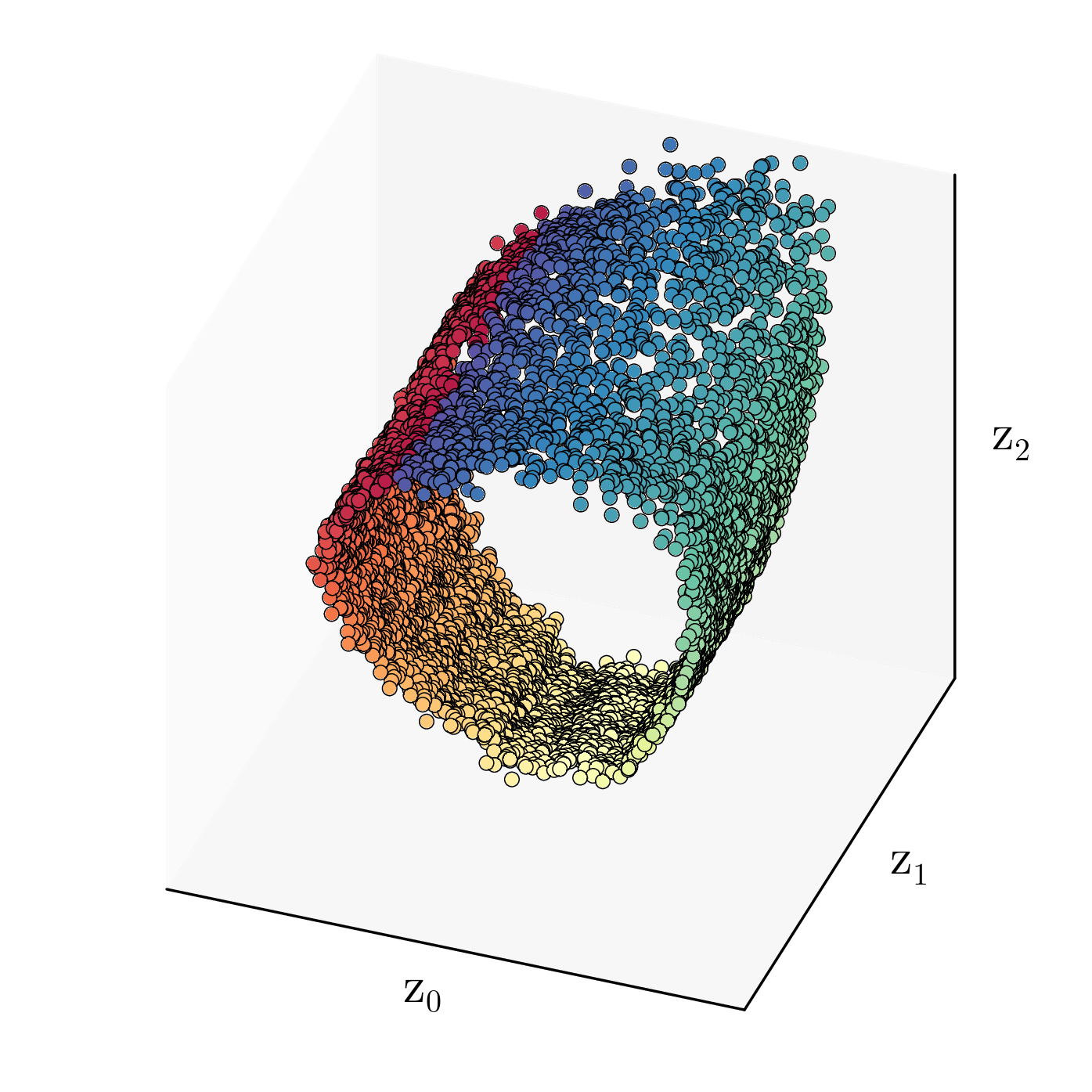}&\includegraphics[width=.2\linewidth]{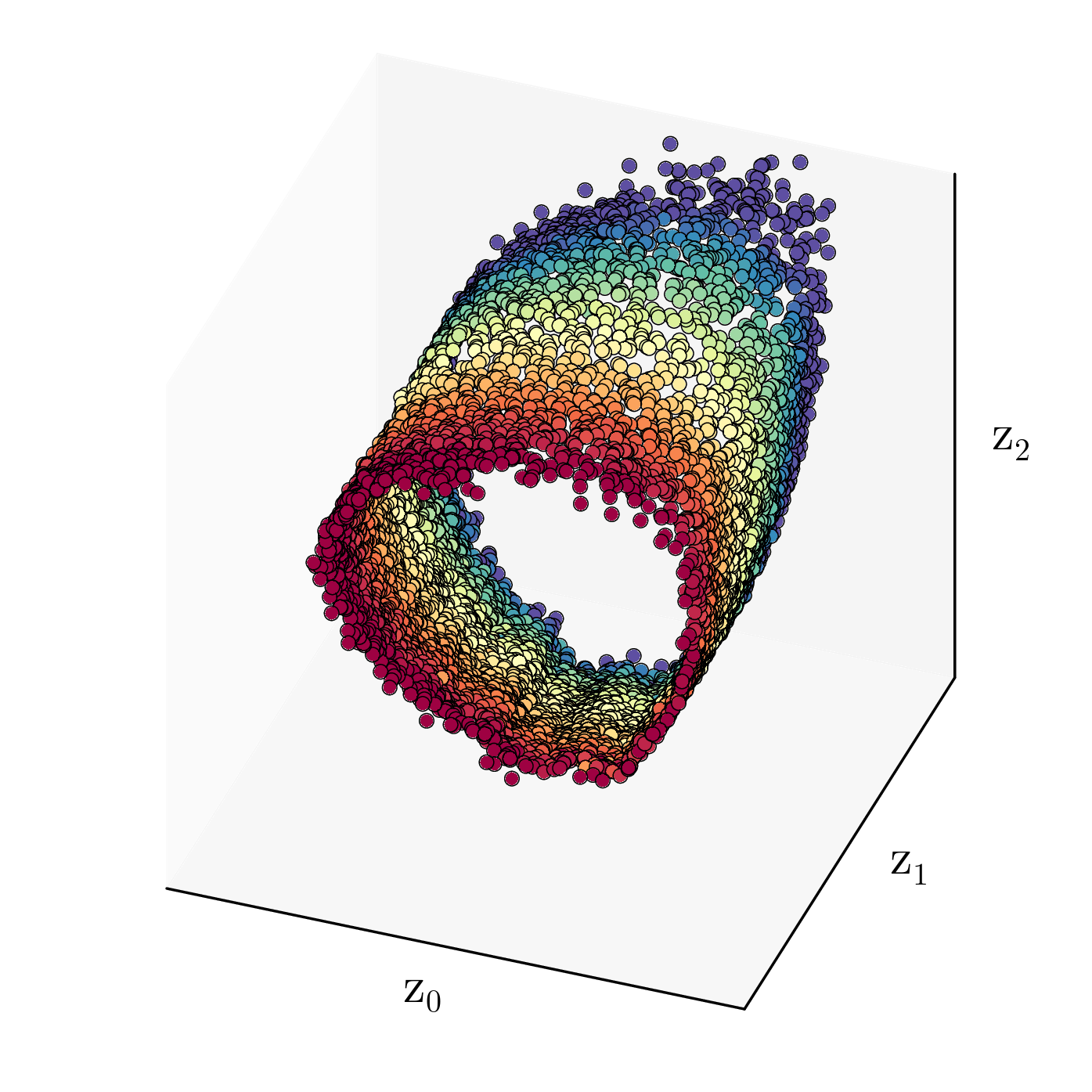}\\
				\includegraphics[width=.2\linewidth]{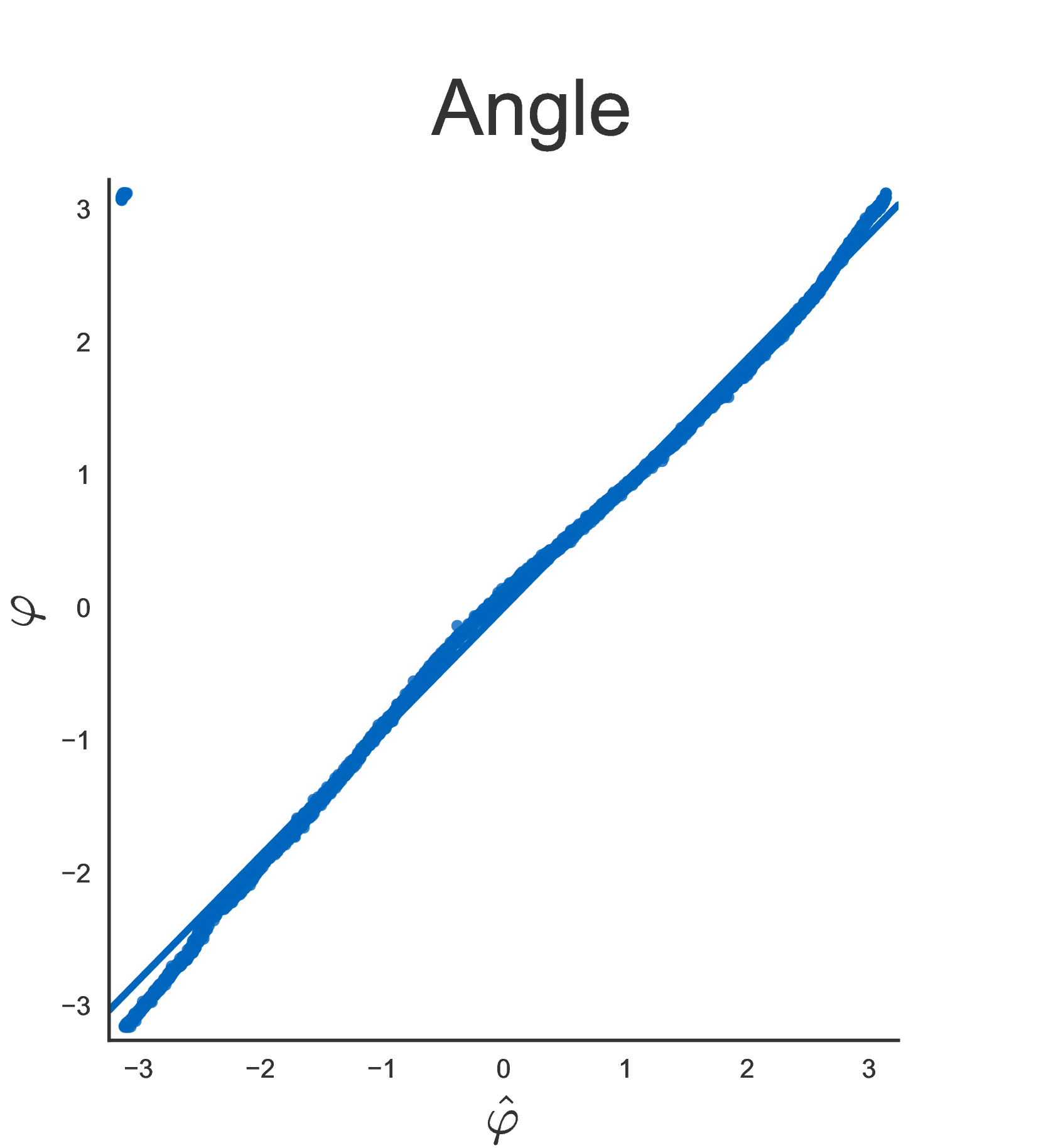}&\includegraphics[width=.2\linewidth]{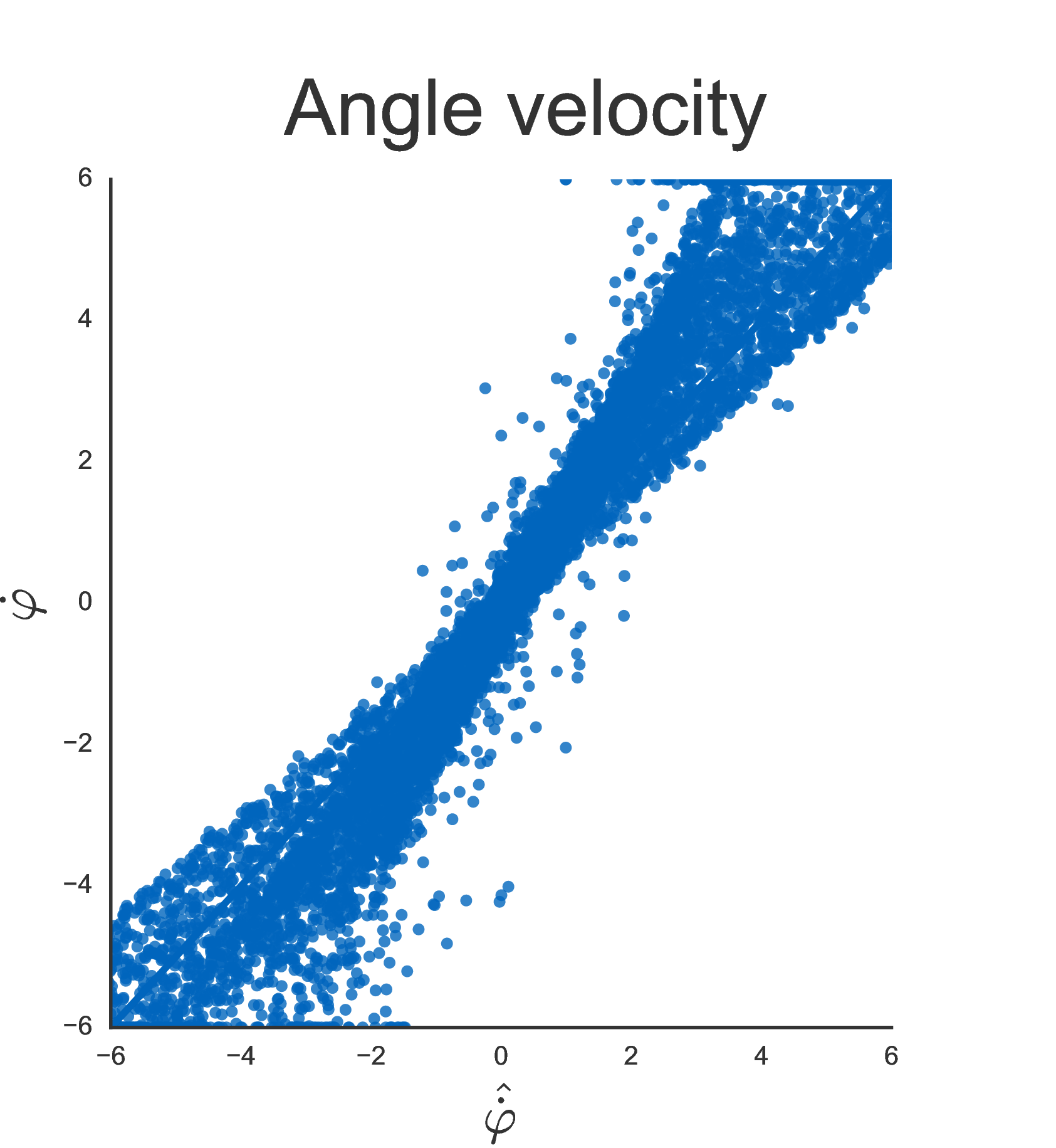}
			\end{tabular}
			\label{fig:dvbf}
		}&
		\subfloat[DKF]{
			\begin{tabular}{cc}
				\includegraphics[width=.2\linewidth]{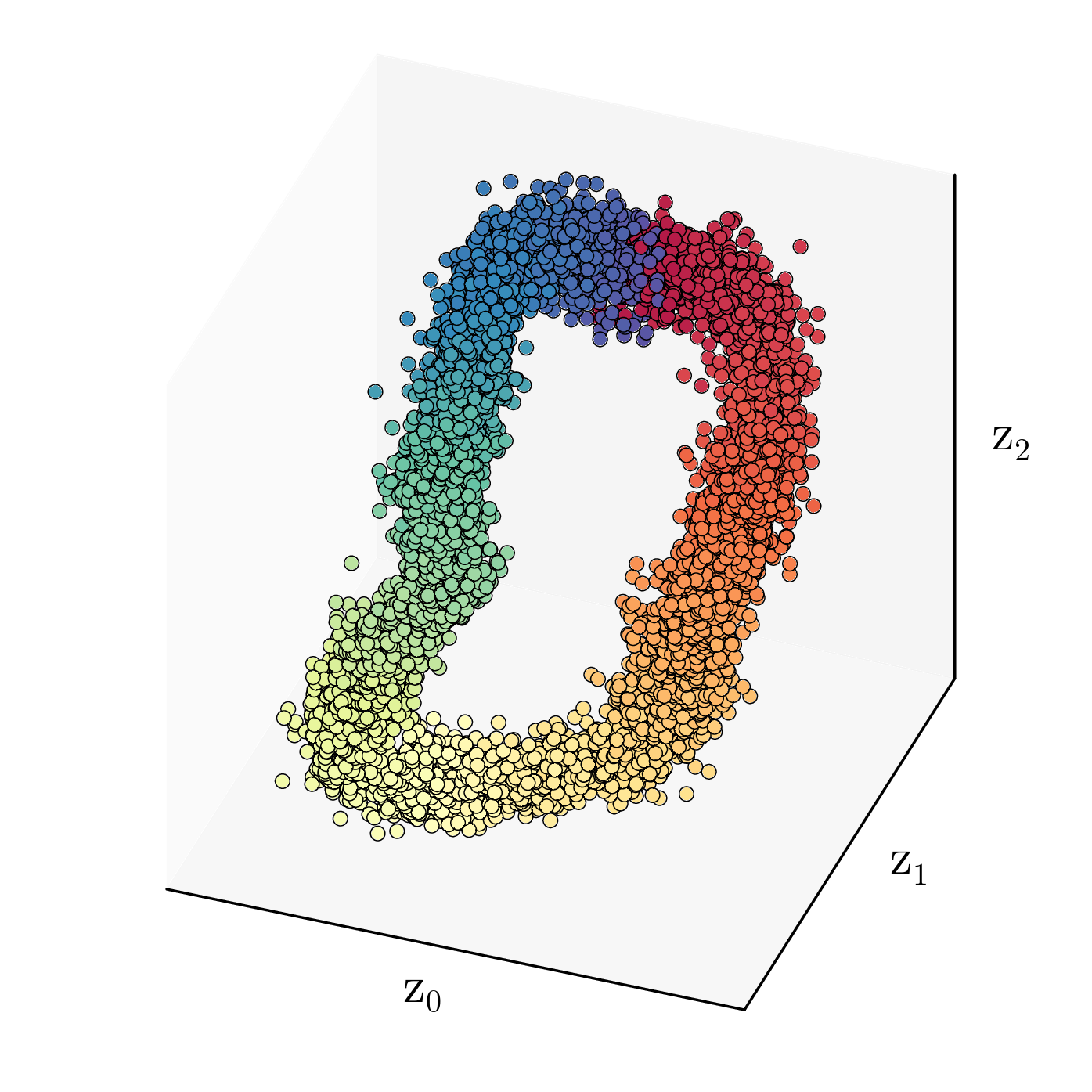}&\includegraphics[width=.2\linewidth]{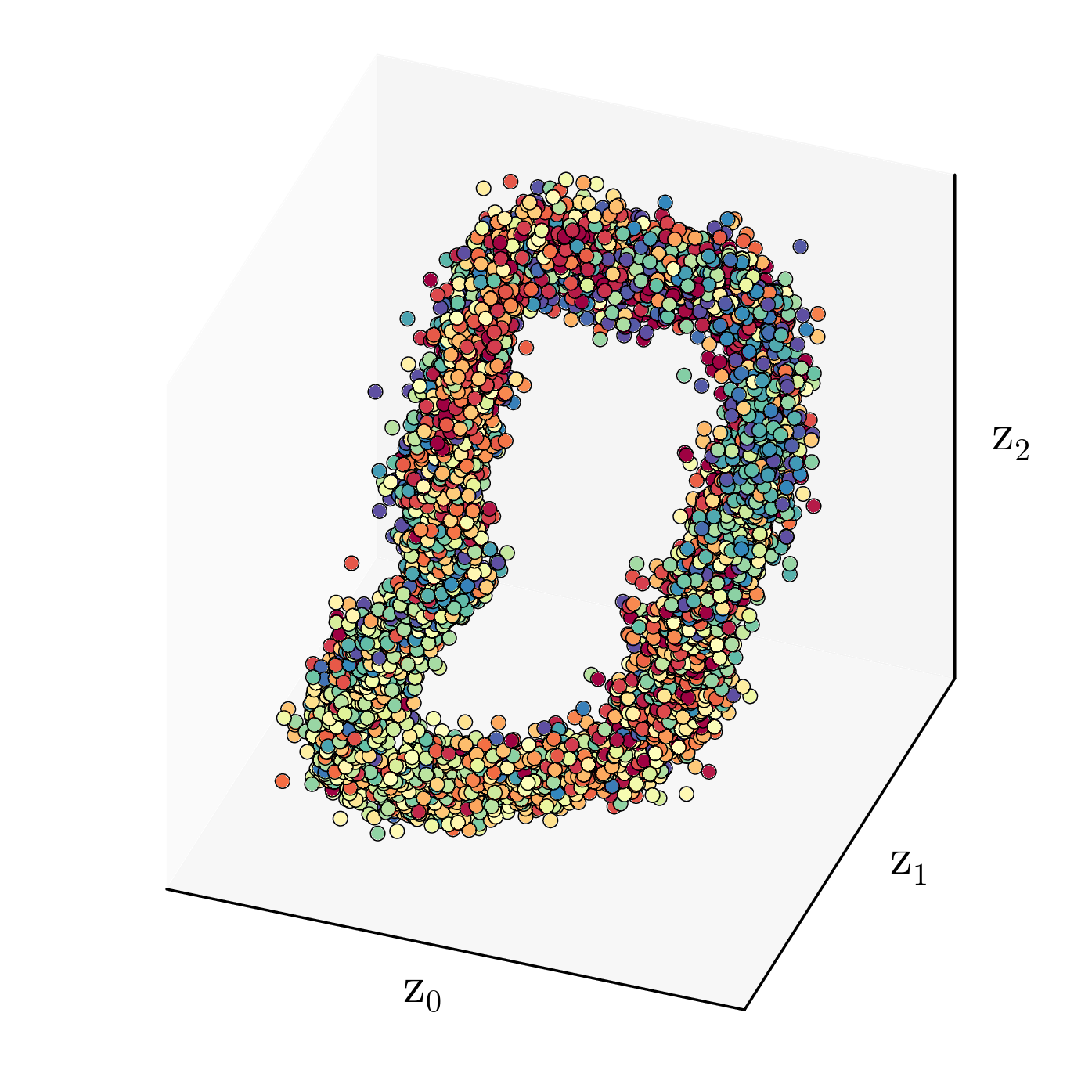}\\
				\includegraphics[width=.2\linewidth]{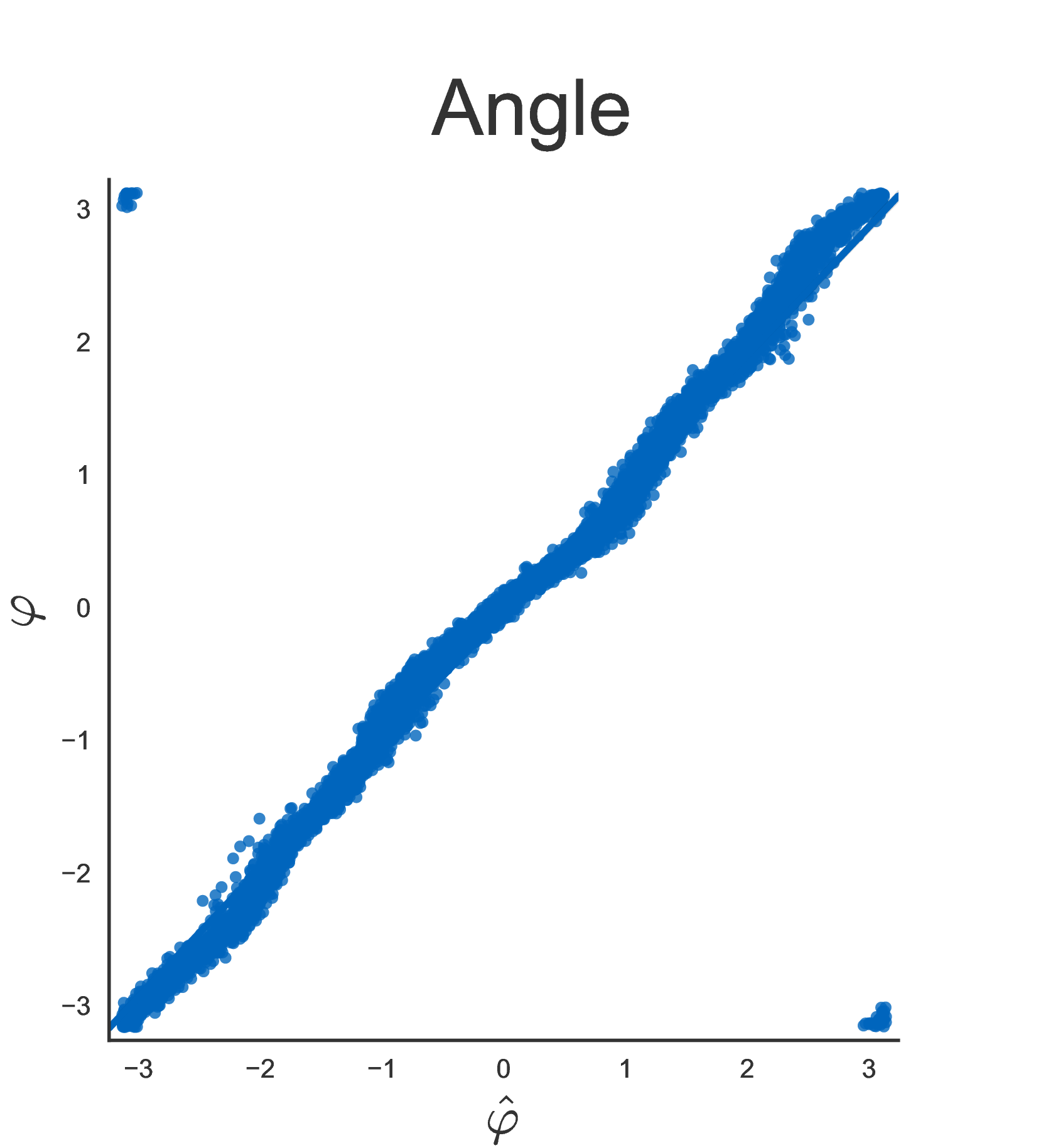}&\includegraphics[width=.2\linewidth]{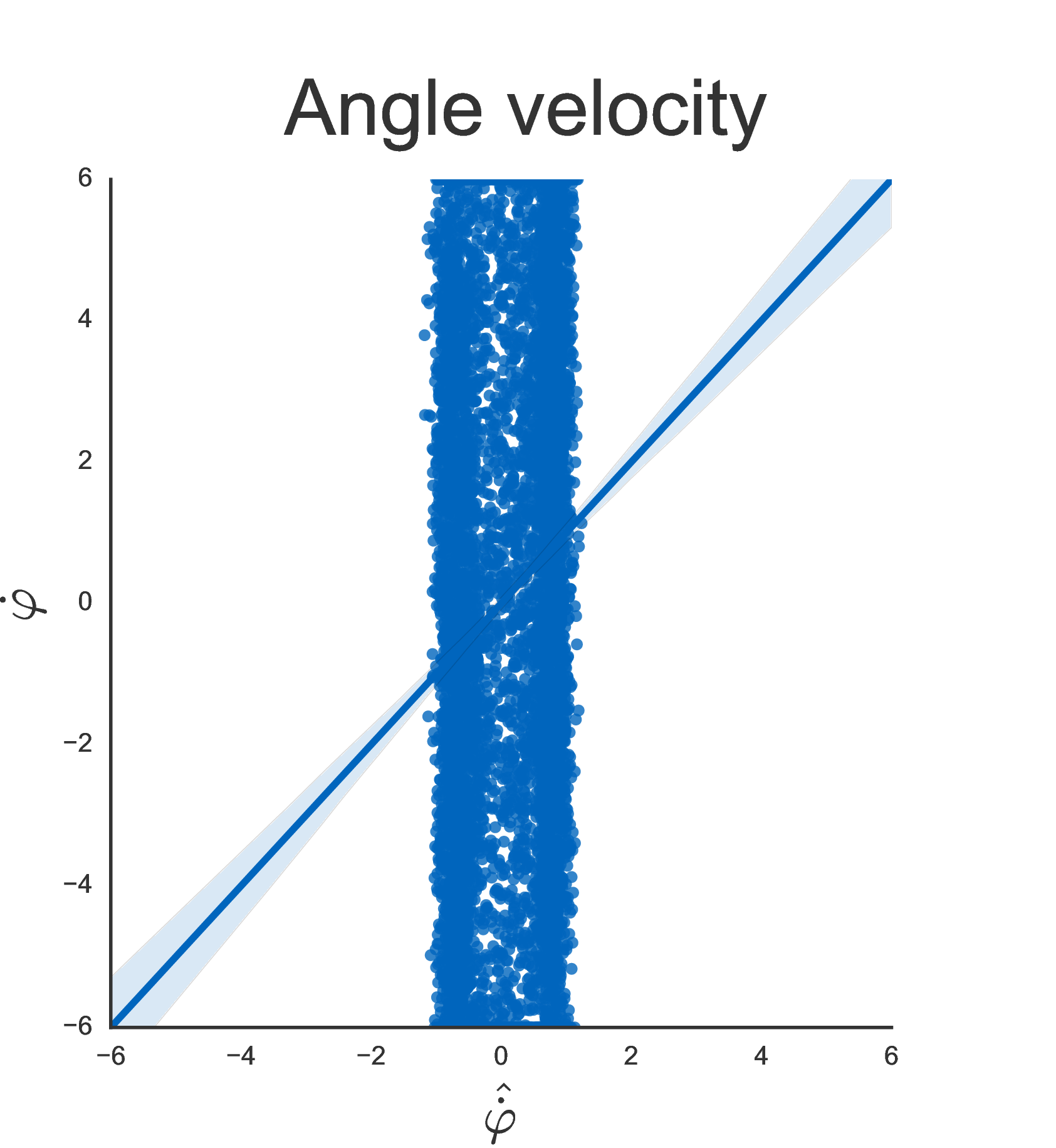}
			\end{tabular}
			\label{fig:dkf}
		}
	\end{tabular}
	\caption{(a) Our DVBF-LL model trained on pendulum image sequences. The upper plots show the latent space with coloring according to the ground truth with angles on the left and angular velocities on the right. The lower plots show regression results for predicting ground truth from the latent representation. The latent space plots show clearly that all information for representing the full state of a pendulum is encoded in each latent state. (b) DKF from \cite{deepkalman} trained on the same pendulum dataset. The latent space plot shows that DKF fails to learn velocities of the pendulum. It is therefore not able to capture all information for representing the full pendulum state.}
	\label{fig:dvbf_vs_dkf}
\end{figure}

In order to test our algorithm on truly non-Markovian observations of a dynamical system, we simulated a dynamic torque-controlled pendulum governed by the differential equation
\eq{{m l^2}\ddot{\varphi}(t) = {- \mu \dot{\varphi}(t) + mgl \sin \varphi(t) + u(t)},}
$m=l=1, \mu=0.5, g=9.81$, via numerical integration, and then converted the ground-truth angle $\varphi$ into an image observation in $\mcX$. The one-dimensional control corresponds to angle acceleration (which is proportional to joint torque). Angle and angular velocity fully describe the system.

\Cref{fig:dvbf_vs_dkf} shows the latent spaces for identical input data learned by DVBF-LL and DKF, respectively, colored with the ground truth in the top row. It should be noted that latent \emph{samples} are shown, \emph{not} means of posterior distributions. The state-space model was allowed to use three latent dimensions. As we can see in \cref{fig:dvbf}, DVBF-LL learned a two-dimensional manifold embedding, i.e., it encoded the angle in polar coordinates (thus circumventing the discontinuity of angles modulo $2\pi$). The bottom row shows ordinary least-squares regressions (OLS) underlining the performance: there exists a high correlation between latent states and ground-truth angle and angular velocity for DVBF-LL. On the contrary, \cref{fig:dkf} verifies our prediction that DKF is equally capable of learning the angle, but extracts little to no information on angular velocity.

\begin{table}
	\caption{Results for pendulum OLS regressions of all latent states on respective dependent variable.}
	\label{tab:regression}
	\centering
	\begin{tabular}{cc}
		\pbox{2cm}{\vspace{2\baselineskip} Dependent\\ ground truth \\variable}&\begin{tabular}{cccccc}
			&\multicolumn{2}{c}{DVBF-LL}&&\multicolumn{2}{c}{DKF}\\
			&Log-Likelihood&$R^2$&\hspace{.2cm}&Log-Likelihood&$R^2$\\\hline
			$\sin(\varphi)$&3990.8&0.961&&1737.6&0.929\\
			$\cos(\varphi)$&7231.1&0.982&&6614.2&0.979\\
			$\dot{\varphi}$&$-$11139&0.916&&$-$20289&0.035
		\end{tabular}
	\end{tabular}
\end{table}
The OLS regression results shown in \cref{tab:regression} validate this observation.\footnote{Linear regression is a natural choice: after transforming the ground truth to polar coordinates, an affine transformation should be a good fit for predicting ground truth from latent states. We also tried nonlinear regression with vanilla neural networks. While not being shown here, the results underlined the same conclusion.} Predicting $\sin(\varphi)$ and $\cos(\varphi)$, i.e., polar coordinates of the ground-truth angle $\varphi$, works almost equally well for DVBF-LL and DKF, with DVBF-LL slightly outperforming DKF. For predicting the ground truth velocity $\dot{\varphi}$, DVBF-LL shows remarkable performance. DKF, instead, contains hardly any information, resulting in a very low goodness-of-fit score of $R^2 = 0.035$.

\begin{figure}\centering
	\subfloat[Generative latent walk.]{\includegraphics[width=.4\linewidth]{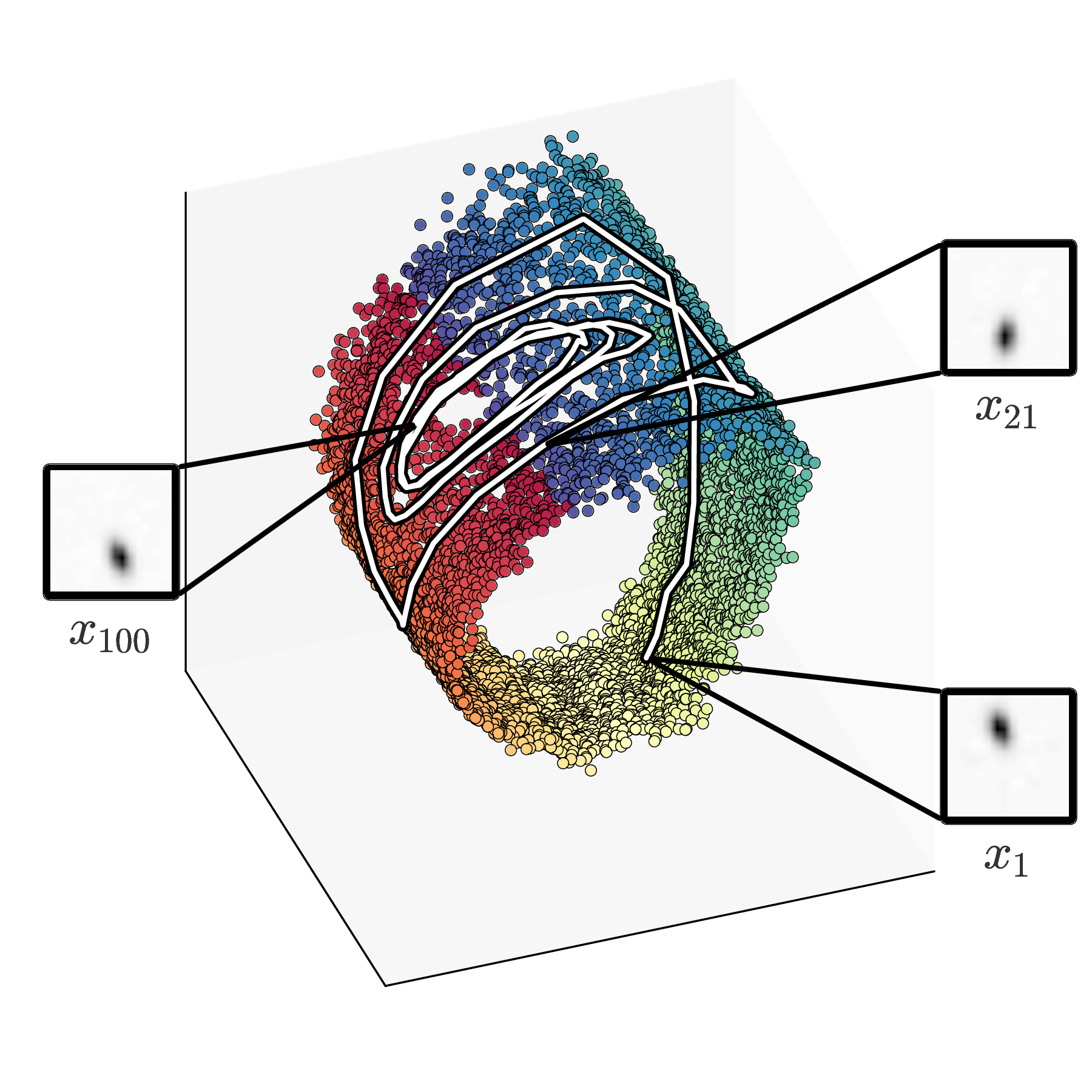}}\qquad
	\subfloat[Reconstructive latent walk.]{\includegraphics[width=.4\linewidth]{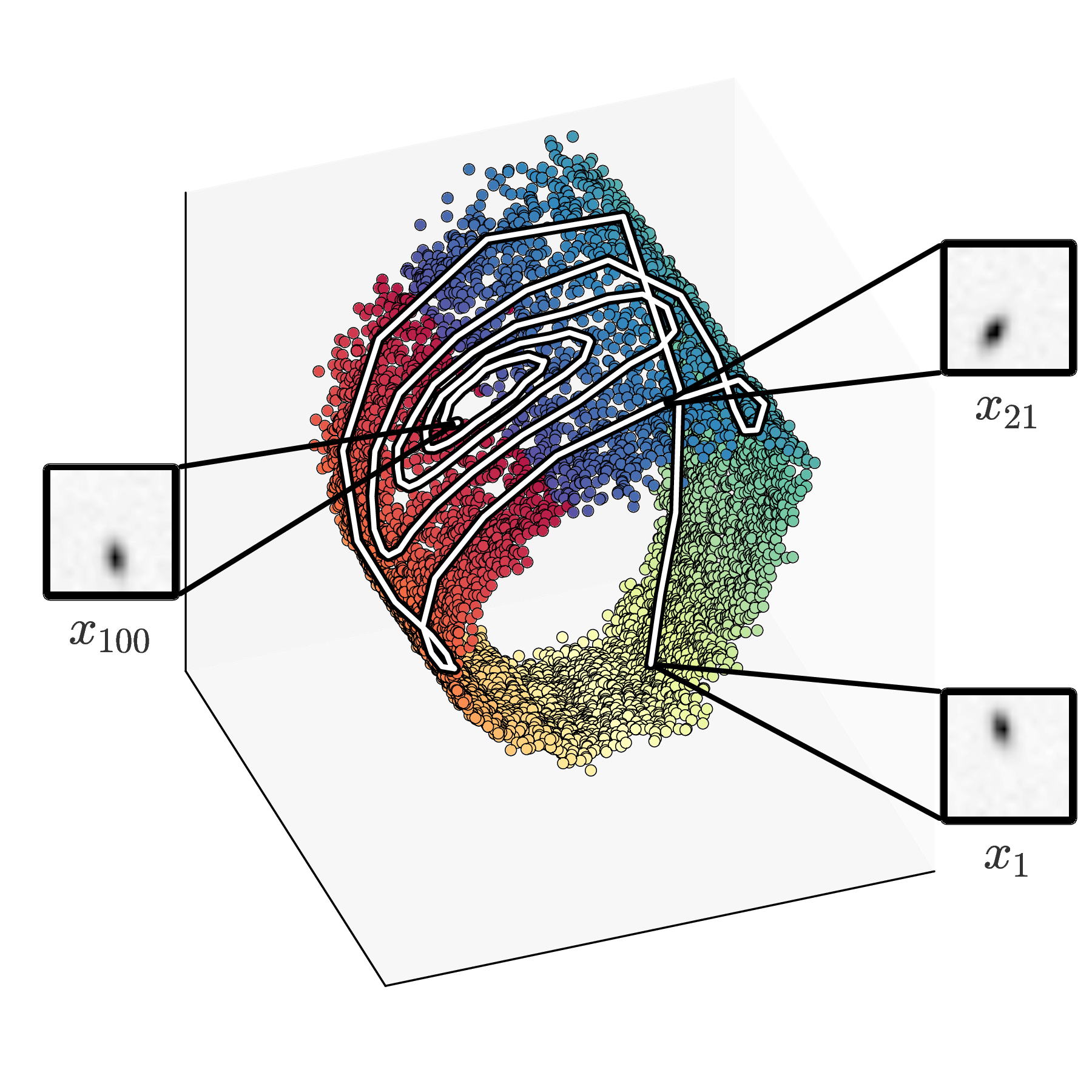}}\\
	\subfloat[Ground truth (top), reconstructions (middle), generative samples (bottom) from identical initial latent state.]{\includegraphics[width=\linewidth]{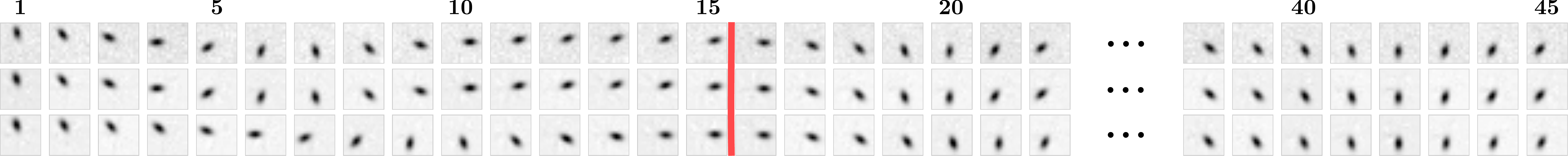}}
	\caption{(a) Latent space walk in generative mode. (b) Latent space walk in filtering mode. (c)~Ground truth and samples from recognition and generative model. The reconstruction sampling has access to observation sequence and performs filtering. The generative samples only get access to the observations once for creating the initial state while all subsequent samples are predicted from this single initial state. The red bar indicates the length of training sequences. Samples beyond show the generalization capabilities for sequences longer than during training. The complete sequence can be found in the Appendix in \cref{fig:samples_dvbf}.}
	\label{fig:pendulum_walks}
\end{figure}

\Cref{fig:pendulum_walks} shows that the strong relation between ground truth and latent state is beneficial for generative sampling. All plots show 100 time steps of a pendulum starting from the exact same latent state and not being actuated. The top row plots show a purely generative walk in the latent space on the left, and a walk in latent space that is corrected by filtering observations on the right. We can see that both follow a similar trajectory to an attractor. The generative model is more prone to noise when approaching the attractor.

The bottom plot shows the first 45 steps of the corresponding observations (top row), reconstructions (middle row), and generative samples (without correcting from observations). Interestingly, DVBF works very well even though the sequence is much longer than all training sequences (indicated by the red line).

\begin{table}
	\caption{Average test set objective function values for pendulum experiment.}
	\label{tab:elbo}
	\centering
	\begin{tabular}{cccccc}
		&Lower Bound&$=$&Reconstruction Error&$-$&KL divergence\\\hline
		DVBF-LL& 798.56 && 802.06 && 3.50\\
		DKF & 784.70 && 788.58 && 3.88
	\end{tabular}
\end{table}

Table~\eqref{tab:elbo} shows values of the lower bound to the marginal data likelihood (for DVBF-LL, this corresponds to \cref{eq:loss_VIF}). We see that DVBF-LL outperforms DKF in terms of compression, but only with a slight margin, which does not reflect the better generative sampling as \cite{theis2015note} argue.

\subsection{Bouncing Ball}
The bouncing ball experiment features a ball rolling within a bounding box in a plane. The system has a two-dimensional control input, added to the directed velocity of the ball. If the ball hits the wall, it bounces off, so that the true dynamics are highly dependent on the current position and velocity of the ball. The system's state is four-dimensional, two dimensions each for position and velocity.

\begin{figure}
	\subfloat[Latent walk of bouncing ball.]{\includegraphics[width=.37\linewidth]{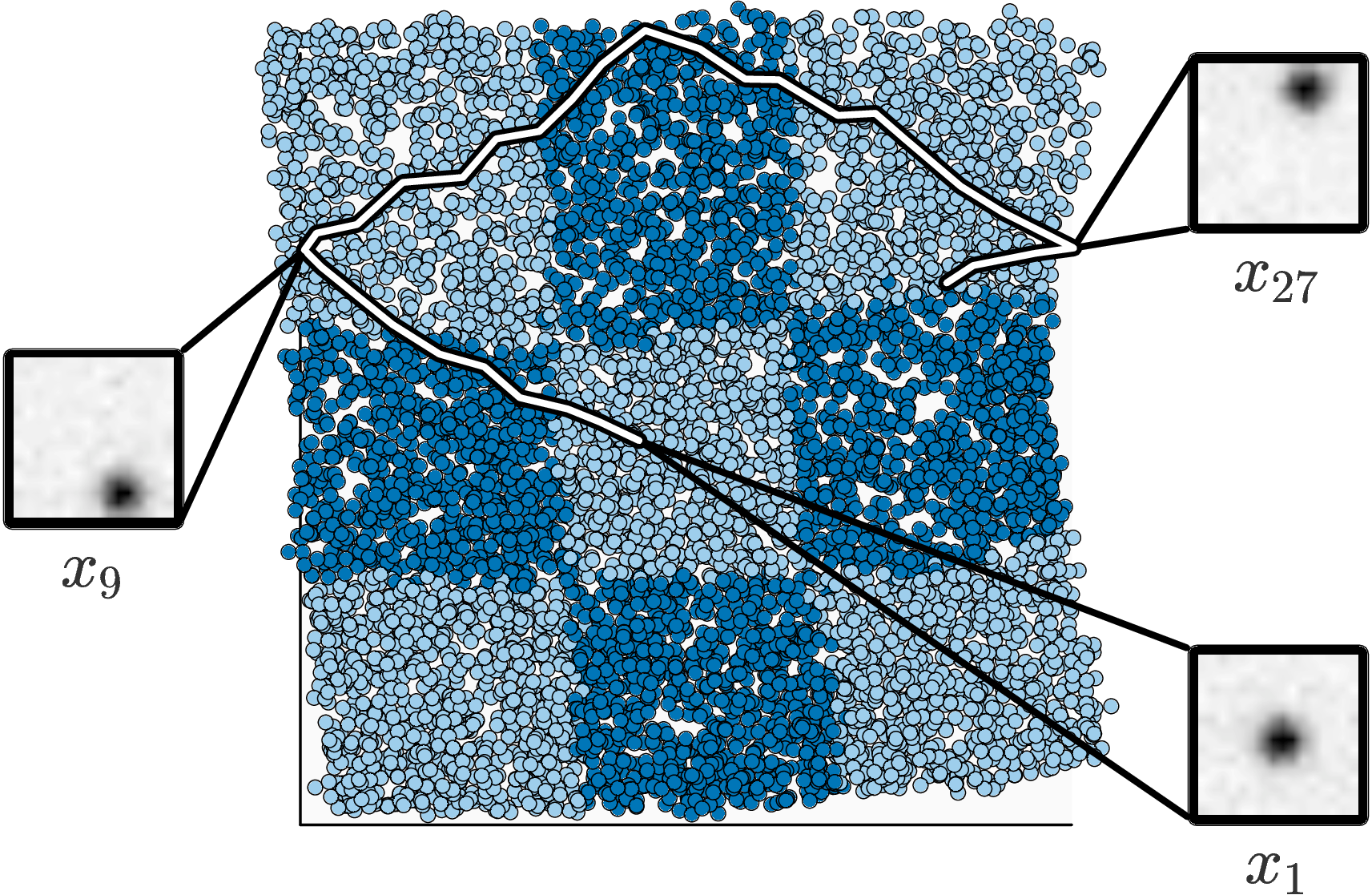}}\qquad
	\subfloat[Latent space velocities.]{
		\includegraphics[width=.27\linewidth]{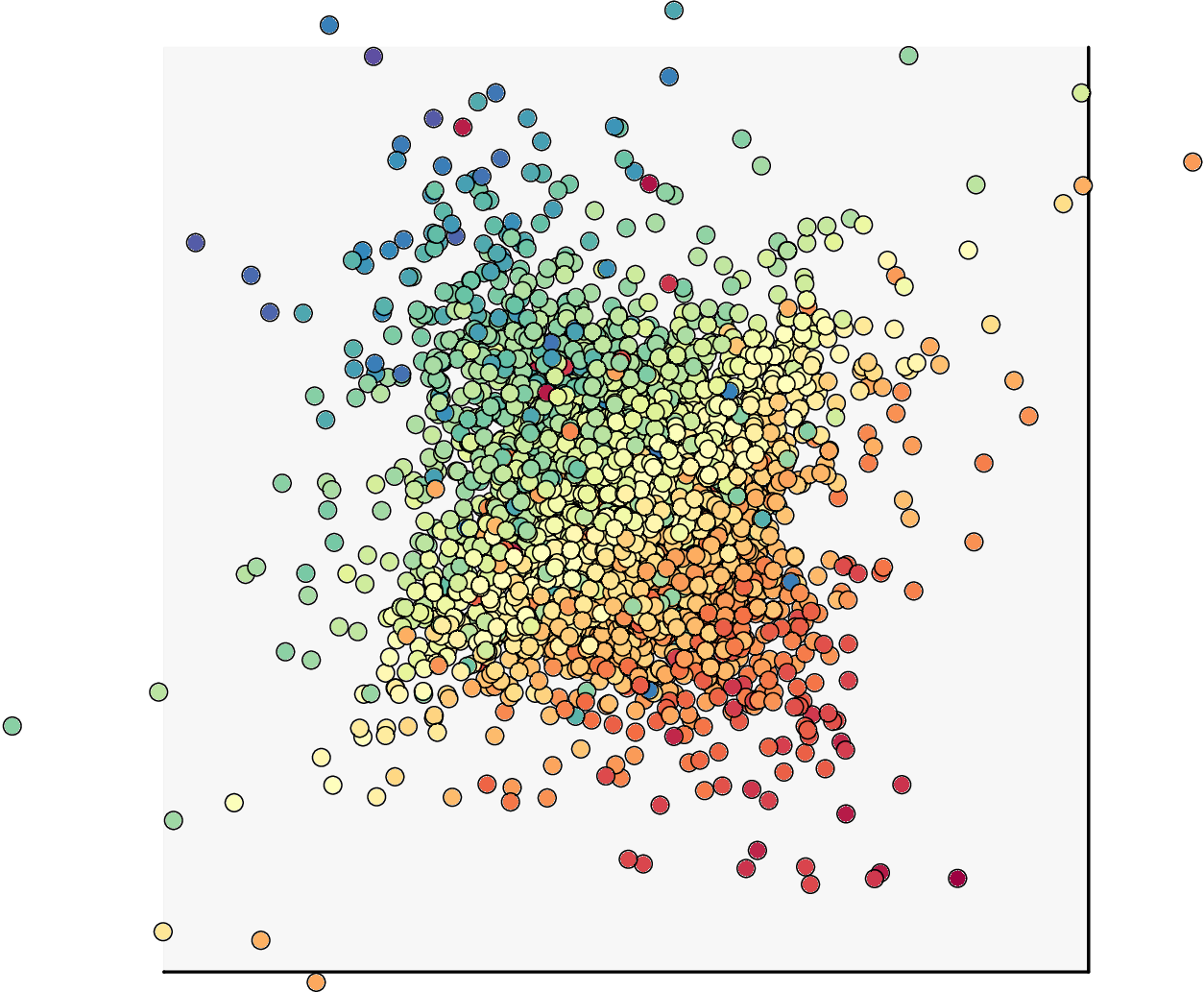}
		\includegraphics[width=.27\linewidth]{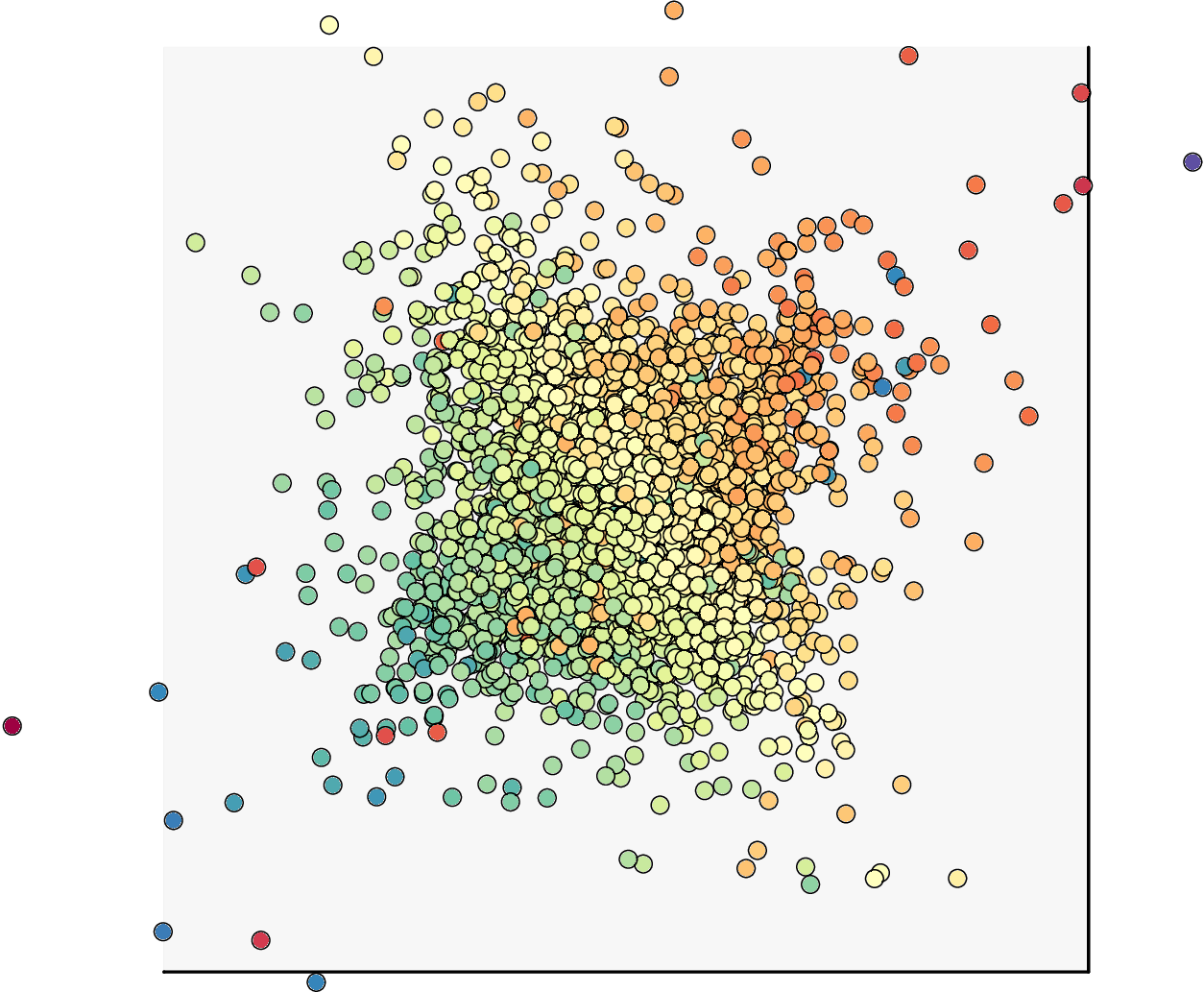}
	}
	\caption{(a) Two dimensions of 4D bouncing ball latent space. Ground truth x and y coordinates are combined into a regular 3$\times$3 checkerboard coloring. This checkerboard is correctly extracted by the embedding. (b) Remaining two latent dimensions. Same latent samples, colored with ball velocities in x and y direction (left and right image, respectively). The smooth, perpendicular coloring indicates that the ground truth value is stored in the latent dimension.}
	\label{fig:bouncing_ball}
\end{figure}

Consequently, we use a DVBF-LL with four latent dimensions. \Cref{fig:bouncing_ball} shows that DVBF again captures the entire system dynamics in the latent space. The checkerboard is quite a remarkable result: the ground truth position of the ball lies within the 2D unit square, the bounding box. In order to visualize how ground truth reappears in the learned latent states, we show the warping of the ground truth bounding box into the latent space. To this end, we partitioned (discretized) the ground truth unit square into a regular 3x3 checkerboard with respective coloring. We observed that DVBF learned to extract the 2D position from the 256 pixels, and aligned them in two dimensions of the latent space in strong correspondence to the physical system. The algorithm does the exact same pixel-to-2D inference that a human observer automatically does when looking at the image.

\subsection{Two Bouncing Balls}

\begin{figure}\centering
    \includegraphics[width=\linewidth]{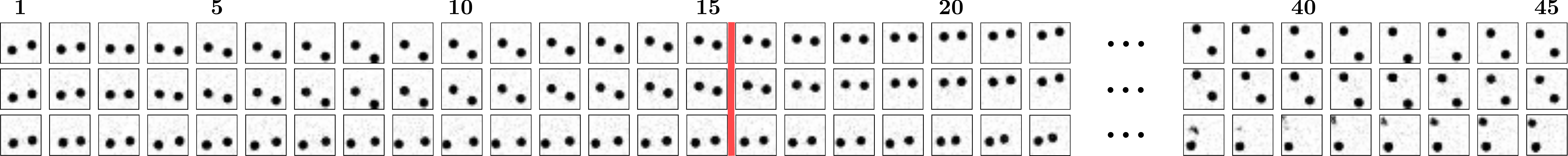}
    \caption{Ground truth (top), reconstructions (middle), generative samples (bottom) from identical initial latent state for the two bouncing balls experiment. Red bar indicates length of training sequences.}
    \label{fig:tbb_results}
\end{figure}

Another more complex environment\footnote{We used the script attached to \cite{AISTATS07_SutskeverH} for generating our datasets.} features two balls in a bounding box.
We used a 10-dimensional latent space to fully capture the position and velocity information of the balls.
Reconstruction and generative samples are shown in \cref{fig:tbb_results}.
Same as in the pendulum example we get a generative model with stable predictions beyond training data sequence length.

\section{Conclusion}

We have proposed Deep Variational Bayes Filters (DVBF), a new method to learn state space models from raw non-Markovian sequence data. 
DVBFs perform latent dynamic system identification, and subsequently overcome intractable inference.
As DVBFs make use of stochastic gradient variational Bayes they naturally scale to large data sets.
In a series of vision-based experiments we demonstrated that latent states can be recovered which identify the underlying physical quantities. The generative model showed stable long-term predictions far beyond the sequence length used during training.

\subsubsection*{Acknowledgements}

Part of this work was conducted at Chair of Robotics and Embedded Systems, Department of Informatics,
Technische Universit\"at M\"unchen, Germany, and supported by the TACMAN project, EC Grant agreement no.\ 610967, within the FP7 framework programme.

We would like to thank Jost Tobias Springenberg, Adam Kosiorek, Moritz M\"unst, and anonymous reviewers for valuable input.


\bibliography{Bibliography}

\begin{thebibliography}{25}
\providecommand{\natexlab}[1]{#1}
\providecommand{\url}[1]{\texttt{#1}}
\expandafter\ifx\csname urlstyle\endcsname\relax
  \providecommand{\doi}[1]{doi: #1}\else
  \providecommand{\doi}{doi: \begingroup \urlstyle{rm}\Url}\fi

\bibitem[Bayer \& Osendorfer(2014)Bayer and Osendorfer]{storn2014}
Justin Bayer and Christian Osendorfer.
\newblock Learning stochastic recurrent networks.
\newblock \emph{arXiv preprint arXiv:1411.7610}, 2014.

\bibitem[Blundell et~al.(2015)Blundell, Cornebise, Kavukcuoglu, and
  Wierstra]{weightuncertainty}
Charles Blundell, Julien Cornebise, Koray Kavukcuoglu, and Daan Wierstra.
\newblock Weight uncertainty in neural networks.
\newblock \emph{arXiv preprint arXiv:1505.05424}, 2015.

\bibitem[Bottou(2010)]{bottou2010large}
L{\'e}on Bottou.
\newblock Large-scale machine learning with stochastic gradient descent.
\newblock In \emph{Proceedings of COMPSTAT'2010}, pp.\  177--186. Springer,
  2010.

\bibitem[Chung et~al.(2015)Chung, Kastner, Dinh, Goel, Courville, and
  Bengio]{vrnn2015}
Junyoung Chung, Kyle Kastner, Laurent Dinh, Kratarth Goel, Aaron~C. Courville,
  and Yoshua Bengio.
\newblock A recurrent latent variable model for sequential data.
\newblock \emph{CoRR}, abs/1506.02216, 2015.
\newblock URL \url{http://arxiv.org/abs/1506.02216}.

\bibitem[Deisenroth \& Rasmussen(2011)Deisenroth and
  Rasmussen]{deisenroth2011pilco}
Marc Deisenroth and Carl~E Rasmussen.
\newblock Pilco: A model-based and data-efficient approach to policy search.
\newblock In \emph{Proceedings of the 28th International Conference on machine
  learning (ICML-11)}, pp.\  465--472, 2011.

\bibitem[Ghahramani \& Hinton(1996)Ghahramani and
  Hinton]{ghahramani1996parameter}
Zoubin Ghahramani and Geoffrey~E Hinton.
\newblock Parameter estimation for linear dynamical systems.
\newblock Technical report, Technical Report CRG-TR-96-2, University of
  Toronto, Dept. of Computer Science, 1996.

\bibitem[Ghahramani \& Hinton(2000)Ghahramani and Hinton]{variationalswitching}
Zoubin Ghahramani and Geoffrey~E Hinton.
\newblock Variational learning for switching state-space models.
\newblock \emph{Neural computation}, 12\penalty0 (4):\penalty0 831--864, 2000.

\bibitem[Graves(2013)]{graves2013generating}
Alex Graves.
\newblock Generating sequences with recurrent neural networks.
\newblock \emph{arXiv preprint arXiv:1308.0850}, 2013.

\bibitem[Hinton \& Van~Camp(1993)Hinton and Van~Camp]{hinton1993keeping}
Geoffrey~E Hinton and Drew Van~Camp.
\newblock Keeping the neural networks simple by minimizing the description
  length of the weights.
\newblock In \emph{Proceedings of the sixth annual conference on Computational
  learning theory}, pp.\  5--13. ACM, 1993.

\bibitem[Honkela et~al.(2010)Honkela, Raiko, Kuusela, Tornio, and
  Karhunen]{honkela2010approximate}
Antti Honkela, Tapani Raiko, Mikael Kuusela, Matti Tornio, and Juha Karhunen.
\newblock Approximate riemannian conjugate gradient learning for fixed-form
  variational bayes.
\newblock \emph{Journal of Machine Learning Research}, 11\penalty0
  (Nov):\penalty0 3235--3268, 2010.

\bibitem[Johnson et~al.(2016)Johnson, Duvenaud, Wiltschko, Datta, and
  Adams]{svae}
Matthew~J Johnson, David Duvenaud, Alexander~B Wiltschko, Sandeep~R Datta, and
  Ryan~P Adams.
\newblock Structured {VAE}s: Composing probabilistic graphical models and
  variational autoencoders.
\newblock \emph{arXiv preprint arXiv:1603.06277}, 2016.

\bibitem[Julier \& Uhlmann(1997)Julier and Uhlmann]{julier1997new}
Simon~J Julier and Jeffrey~K Uhlmann.
\newblock New extension of the kalman filter to nonlinear systems.
\newblock In \emph{AeroSense'97}, pp.\  182--193. International Society for
  Optics and Photonics, 1997.

\bibitem[Kalman \& Bucy(1961)Kalman and Bucy]{kalman1961new}
Rudolph~E Kalman and Richard~S Bucy.
\newblock New results in linear filtering and prediction theory.
\newblock \emph{Journal of basic engineering}, 83\penalty0 (1):\penalty0
  95--108, 1961.

\bibitem[Kingma \& Welling(2013)Kingma and Welling]{vae2013}
Diederik~P Kingma and Max Welling.
\newblock Auto-encoding variational bayes.
\newblock \emph{arXiv preprint arXiv:1312.6114}, 2013.

\bibitem[Ko \& Fox(2011)Ko and Fox]{ko2011learning}
Jonathan Ko and Dieter Fox.
\newblock Learning gp-bayesfilters via gaussian process latent variable models.
\newblock \emph{Autonomous Robots}, 30\penalty0 (1):\penalty0 3--23, 2011.

\bibitem[Krishnan et~al.(2015)Krishnan, Shalit, and Sontag]{deepkalman}
Rahul~G Krishnan, Uri Shalit, and David Sontag.
\newblock Deep {K}alman filters.
\newblock \emph{arXiv preprint arXiv:1511.05121}, 2015.

\bibitem[Mandt et~al.(2016)Mandt, McInerney, Abrol, Ranganath, and
  Blei]{variationaltempering}
Stephan Mandt, James McInerney, Farhan Abrol, Rajesh Ranganath, and David Blei.
\newblock Variational tempering.
\newblock In \emph{Proceedings of the 19th International Conference on
  Artificial Intelligence and Statistics}, pp.\  704--712, 2016.

\bibitem[McGoff et~al.(2015)McGoff, Mukherjee, Pillai,
  et~al.]{reviewsystemstheory}
Kevin McGoff, Sayan Mukherjee, Natesh Pillai, et~al.
\newblock Statistical inference for dynamical systems: A review.
\newblock \emph{Statistics Surveys}, 9:\penalty0 209--252, 2015.

\bibitem[Rezende et~al.(2014)Rezende, Mohamed, and Wierstra]{dlgm2014}
Danilo~J. Rezende, Shakir Mohamed, and Daan Wierstra.
\newblock Stochastic backpropagation and approximate inference in deep
  generative models.
\newblock In Tony Jebara and Eric~P. Xing (eds.), \emph{Proceedings of the 31st
  International Conference on Machine Learning (ICML-14)}, pp.\  1278--1286.
  JMLR Workshop and Conference Proceedings, 2014.
\newblock URL \url{http://jmlr.org/proceedings/papers/v32/rezende14.pdf}.

\bibitem[Rezende \& Mohamed(2015)Rezende and Mohamed]{normflow}
Danilo~Jimenez Rezende and Shakir Mohamed.
\newblock Variational inference with normalizing flows.
\newblock \emph{arXiv preprint arXiv:1505.05770}, 2015.

\bibitem[Sutskever \& Hinton(2007)Sutskever and Hinton]{AISTATS07_SutskeverH}
Ilya Sutskever and Geoffrey~E. Hinton.
\newblock Learning multilevel distributed representations for high-dimensional
  sequences.
\newblock In Marina Meila and Xiaotong Shen (eds.), \emph{Proceedings of the
  Eleventh International Conference on Artificial Intelligence and Statistics
  (AISTATS-07)}, volume~2, pp.\  548--555. Journal of Machine Learning Research
  - Proceedings Track, 2007.
\newblock URL
  \url{http://jmlr.csail.mit.edu/proceedings/papers/v2/sutskever07a/sutskever07a.pdf}.

\bibitem[Sutton(1996)]{sutton1996model}
Leonid Kuvayev~Rich Sutton.
\newblock Model-based reinforcement learning with an approximate, learned
  model.
\newblock In \emph{Proceedings of the ninth Yale workshop on adaptive and
  learning systems}, pp.\  101--105, 1996.

\bibitem[Theis et~al.(2015)Theis, Oord, and Bethge]{theis2015note}
Lucas Theis, A{\"a}ron van~den Oord, and Matthias Bethge.
\newblock A note on the evaluation of generative models.
\newblock \emph{arXiv preprint arXiv:1511.01844}, 2015.

\bibitem[Valpola \& Karhunen(2002)Valpola and
  Karhunen]{valpola2002unsupervised}
Harri Valpola and Juha Karhunen.
\newblock An unsupervised ensemble learning method for nonlinear dynamic
  state-space models.
\newblock \emph{Neural computation}, 14\penalty0 (11):\penalty0 2647--2692,
  2002.

\bibitem[Watter et~al.(2015)Watter, Springenberg, Boedecker, and
  Riedmiller]{e2c}
Manuel Watter, Jost Springenberg, Joschka Boedecker, and Martin Riedmiller.
\newblock Embed to control: A locally linear latent dynamics model for control
  from raw images.
\newblock In \emph{Advances in Neural Information Processing Systems}, pp.\
  2728--2736, 2015.

\end{thebibliography}
\bibliographystyle{iclr2017_conference}

\newpage
\appendix

\section{Supplementary to Lower Bound}

\subsection{Annealed KL-Divergence}

We used the analytical solution of the annealed KL-divergence in \cref{eq:loss_VIF_pre_anneal} for optimization:

\eq{
\expectunder{- \ln\qundergiven{\wTs}\phi{\xTs, \uTs} + c_i\ln\p{\wTs}}{\qphi} = \\
c_i \frac{1}{2} \ln(2 \pi \sigma_p^2) - \frac{1}{2} \ln(2 \pi \sigma_q^2) + c_i \frac{\sigma_q^2 + (\mu_q - \mu_p)^2}{2 \sigma_p^2} - \frac{1}{2}
}

\section{Supplementary to Implementation}

\subsection{Experimental setup}
In all our experiments, we use sequences of 15 raw images of the respective system with 16$\times$16 pixels each, i.e., observation space $\mcX\subset\RR^{256}$, as well as control inputs of varying dimension and interpretation depending on the experiment. We used training, validation and test sets with 500 sequences each. Control input sequences were drawn randomly (``motor babbling''). Additional details about the implementation can be found in the published code at \url{https://brml.org/projects/dvbf}.

\subsection{Additional Experiment Plots}

\begin{figure}[h]\centering
    \includegraphics[width=\linewidth]{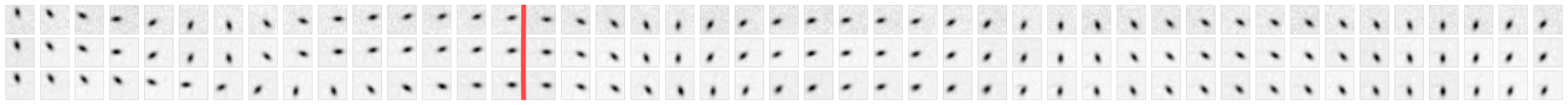}
    \caption{Ground truth and samples from recognition and generative model. Complete version of \cref{fig:pendulum_walks} with all missing samples present.}
    \label{fig:samples_dvbf}
\end{figure}

\subsection{Implementation details for DVBF in Pendulum Experiment}

\begin{itemize}
\item Input: 15 timesteps of $16^2$ observation dimensions and 1 action dimension
\item Latent Space: 3 dimensions
\item Observation Network $p(\bx_t|\bz_t) = \mathcal{N}(\bx_t;\mu(\bz_t), \sigma)$: 128 ReLU + $16^2$ identity output
\item Recognition Model: 128 ReLU + 6 identity output \eq{q(\bw_t|\bz_t, \bx_{t+1}, \bu_t) = \mathcal{N}(\bw_t; \mu, \sigma),\\
(\mu, \sigma) = f(\bz_t, \bx_{t+1}, \bu_t)}\\
\item Transition Network $\boldsymbol{\alpha}_t(\bz_t)$: 16 softmax output
\item Initial Network $\bw_1 \sim p(\xTs)$: Fast Dropout BiRNN with: 128 ReLU + 3 identity output
\item Initial Transition $\bz_1(\bw_1)$: 128 ReLU + 3 identity output
\item Optimizer: adadelta, 0.1 step rate
\item Inverse temperature: $c_0=0.01$, updated every 250th gradient update, $T_A = 10^5$ iterations
\item Batch-size: 500
\end{itemize}

\subsection{Implementation details for DVBF in Bouncing Ball Experiment}

\begin{itemize}
\item Input: 15 timesteps of $16^2$ observation dimensions and 2 action dimension
\item Latent Space: 4 dimensions
\item Observation Network $p(\bx_t|\bz_t) = \mathcal{N}(\bx_t;\mu(\bz_t), \sigma)$: 128 ReLU + $16^2$ identity output
\item Recognition Model: 128 ReLU + 8 identity output \eq{q(\bw_t|\bz_t, \bx_{t+1}, \bu_t) = \mathcal{N}(\bw_t; \mu, \sigma),\\
(\mu, \sigma) = f(\bz_t, \bx_{t+1}, \bu_t)}\\

\item Transition Network $\boldsymbol{\alpha}_t(\bz_t)$: 16 softmax output
\item Initial Network $\bw_1 \sim p(\xTs)$: Fast Dropout BiRNN with: 128 ReLU + 4 identity output
\item Initial Transition $\bz_1(\bw_1)$: 128 ReLU + 4 identity output
\item Optimizer: adadelta, 0.1 step rate
\item Inverse temperature: $c_0=0.01$, updated every 250th gradient update, $T_A = 10^5$ iterations
\item Batch-size: 500
\end{itemize}

\subsection{Implementation details for DVBF in Two Bouncing Balls Experiment}

\begin{itemize}
\item Input: 15 timesteps of $20^2$ observation dimensions and 2000 samples
\item Latent Space: 10 dimensions
\item Observation Network $p(\bx_t|\bz_t) = \mathcal{N}(\bx_t;\mu(\bz_t), \sigma)$: 128 ReLU + $20^2$ sigmoid output
\item Recognition Model: 128 ReLU + 20 identity output \eq{q(\bw_t|\bz_t, \bx_{t+1}, \bu_t) = \mathcal{N}(\bw_t; \mu, \sigma),\\
(\mu, \sigma) = f(\bz_t, \bx_{t+1}, \bu_t)}\\

\item Transition Network $\boldsymbol{\alpha}_t(\bz_t)$: 64 softmax output
\item Initial Network $\bw_1 \sim p(\xTs)$: MLP with: 128 ReLU + 10 identity output
\item Initial Transition $\bz_1(\bw_1)$: 128 ReLU + 10 identity output
\item Optimizer: adam, 0.001 step rate
\item Inverse temperature: $c_0=0.01$, updated every gradient update, $T_A = 2~10^5$ iterations
\item Batch-size: 80
\end{itemize}

\subsection{Implementation details for DKF in Pendulum Experiment}

\begin{itemize}
\item Input: 15 timesteps of $16^2$ observation dimensions and 1 action dimension
\item Latent Space: 3 dimensions
\item Observation Network $p(\bx_t|\bz_t) = \mathcal{N}(\bx_t;\mu(\bz_t), \sigma(\bz_t))$: 128 Sigmoid + 128 Sigmoid + $2~16^2$ identity output
\item Recognition Model: Fast Dropout BiRNN 128 Sigmoid + 128 Sigmoid + 3 identity output

\item Transition Network $p(\bz_t |\bz_{t-1}, \bu_{t-1})$: 128 Sigmoid + 128 Sigmoid + 6 output
\item Optimizer: adam, 0.001 step rate
\item Inverse temperature: $c_0=0.01$, updated every 25th gradient update, $T_A = 2000$ iterations
\item Batch-size: 500
\end{itemize}

\end{document}